\documentclass[lettersize,journal]{IEEEtran}
\usepackage{amsmath,amsfonts}
\usepackage{algorithmic}
\usepackage{algorithm}
\usepackage{array}
\usepackage[caption=false,font=normalsize,labelfont=sf,textfont=sf]{subfig}
\usepackage{textcomp}
\usepackage{stfloats}
\usepackage{url}
\usepackage{verbatim}
\usepackage{graphicx}
\usepackage{cite}
\usepackage{booktabs}
\usepackage{amssymb}
\usepackage{bbding}
\usepackage[numbers,sort&compress]{natbib}

\begin{document}

\title{AR-MOT: Autoregressive Multi-object Tracking}

\author{Lianjie Jia$^\ast$, Yuhan Wu$^\ast$, Binghao Ran$^\ast$, Yifan Wang, Lijun Wang$^\dag$, Huchuan Lu,~\IEEEmembership{Fellow,~IEEE,}
\IEEEcompsocitemizethanks{
\IEEEcompsocthanksitem{$\ast$ Equal contribution, $^\dag$ Corresponding author: Lijun Wang.}
\IEEEcompsocthanksitem {The authors are with Dalian University of Technology, China.}
}
}



\maketitle

\begin{abstract}
As multi-object tracking (MOT) tasks continue to evolve toward more general and multi-modal scenarios, the rigid and task-specific architectures of existing MOT methods increasingly hinder their applicability across diverse tasks and limit flexibility in adapting to new tracking formulations. Most approaches rely on fixed output heads and bespoke tracking pipelines, making them difficult to extend to more complex or instruction-driven tasks. To address these limitations, we propose AR-MOT, a novel autoregressive paradigm that formulates MOT as a sequence generation task within a large language model (LLM) framework. This design enables the model to output structured results through flexible sequence construction, without requiring any task-specific heads. To enhance region-level visual perception, we introduce an Object Tokenizer based on a pretrained detector. To mitigate the misalignment between global and regional features, we propose a Region-Aware Alignment (RAA) module, and to support long-term tracking, we design a Temporal Memory Fusion (TMF) module that caches historical object tokens. AR-MOT offers strong potential for extensibility, as new modalities or instructions can be integrated by simply modifying the output sequence format without altering the model architecture. Extensive experiments on MOT17 and DanceTrack validate the feasibility of our approach, achieving performance comparable to state-of-the-art methods while laying the foundation for more general and flexible MOT systems.
\end{abstract}


\section{Introduction}

\IEEEPARstart{M}{ulti-Object} tracking (MOT) is a fundamental problem in computer vision, which aims to continuously track multiple instances throughout a video sequence. It is widely applied in domains such as intelligent surveillance \cite{ghasemi2015survey}, autonomous driving, and multimedia intelligence.

Current research in MOT follows two prominent paradigms: Tracking-by-Detection (TBD) and Tracking-by-Query (TBQ). TBD methods typically adopt a detection-and-association architecture that first detects objects of interest and then performs complex temporal association. To avoid the explicit association stage, TBQ extends the DETR~\cite{carion2020end} framework by introducing detection and tracking queries, along with a dedicated mechanism for their statement transformation. Continuous tracking is achieved by decoding tracking queries across frames within a unified transformer architecture. While TBD framework is modular and widely adopted, it often involves numerous hand-crafted modules particularly in the association stage. These components limit the scalability and adaptability of the framework across diverse tracking scenarios. Although the TBQ framework eliminates the explicit data association, its pipeline with fixed input-output format limits the model's flexibility, making it difficult to adapt to tasks beyond its original domain.

\IEEEpubidadjcol

\begin{figure}[t]
\centering
\includegraphics[width=\linewidth]{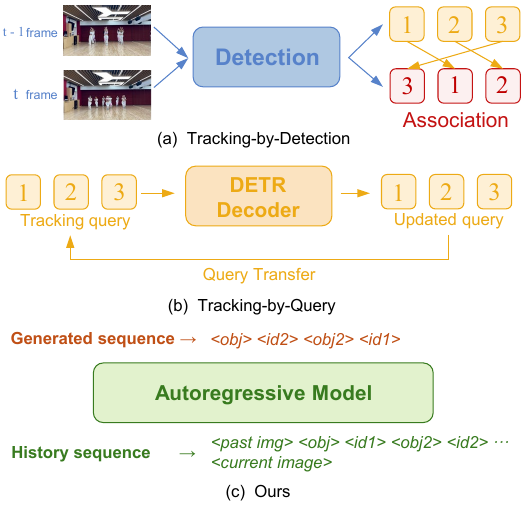}
\caption{\textbf{Comparison of different MOT paradigms.} (a) Tracking by Detection paradiam. (b) Tracking by Query paradiam. (c) Tracking by autoregressively generation.}
\label{fig_1}
\end{figure}

Autoregressive models have been widely applied to vision and multimodal tasks due to their powerful sequence modeling abilities. The success of vision models such as Pix2Seq~\cite{chen2021pix2seq}, OFA~\cite{wang2022ofa}, and Florence-2~\cite{xiao2024florence} validates the effectiveness and adaptability of the autoregressive paradigm in vision tasks. Autoregressive MLLMs, benefiting from training on internet-scale data, have exhibited emergent capabilities to solve general-purpose tasks through next-token prediction. Recently, many MLLM models such as LLaVA~\cite{liu2023visual}, LISA~\cite{lai2024lisa}, and InternVL~\cite{chen2024internvl}, have demonstrated the ability to perform fundamental vision tasks, such as object detection and segmentation in a language-based form. However, existing MLLMs still exhibit notable limitations in addressing the task of MOT, which remains largely unexplored on autoregressive multimodal modeling. MOT is inherently a complex video understanding task that requires not only accurate spatial perception but also robust temporal reasoning to maintain consistent object identities across consecutive frames. Consequently, extending autoregressive MLLMs to support MOT tasks presents substantial challenges.




To address the above challenges and enhance the generality of MOT models, this paper explores a novel approach to multi-object tracking by reformulating the task within an autoregressive framework as illustrated in Fig. \ref{fig_1}. Specifically, we project all MOT-related history information of each frame into MLLM's space to construct structured sequences and prompt the model to predict the corresponding identity for each objects. This design allows the MOT task to be carried out within the LLM framework,  enabling the seamless integration of other modalities such as text, while eliminating the need for task-specific output heads tailored to particular result formats, thereby unlocking significant potential for its integration with and extension to a wide range of other tasks.

Although the above approach has shown initial success in MOT tasks, current MLLMs such as LLaVA and LLaVA-Next~\cite{li2024llava} remain limited in fine-grained perception due to their use of low-resolution image inputs, which hinders accurate localization and discrimination of visually similar or small objects. To compensate for the weakness, we introduce an object tokenizer that performs region-level perception by encoding localized visual features. These features are subsequently projected into the LLM token space, thereby enriching the model's ability of fine-grained perception. This design transforms the MOT task into a question-answering process between the object detector and the MLLM, providing clearer task objectives for the model. This paradigm avoids explicit association for detection results as in TBD methods, and eliminates the complex query state management between tracking and detection queries required by TBQ approaches.

Despite the use of adapter modules for feature alignment, a gap still exists between the representations from the image tokenizer and the object tokenizer. This is primarily due to the fact that the former captures global visual context, whereas the latter encodes fine-grained, region-level details. These inconsistencies can negatively impact the model's ability to maintain coherent and unified visual understanding across different perceptual hierarchies. To effectively bridge this gap, we introduce a novel Region-Aware Alignment (RAA) module, which guides the model to associate and align the visual representations of the same object across the global and local branches. This mechanism enhances the consistency and coherence of cross-branch feature representations, thereby improving the model’s capacity to perform fine-grained, identity-preserving visual reasoning in complex scenarios.

Under the limitations of computational resources and considerations of algorithmic efficiency, it is unreasonable to directly utilize all historical frame information for predicting the current frame. Moreover, due to significant variations in object motion and spatial position, frames that are temporally distant from the current frame may lose their reference value and even negatively impact tracking performance. Therefore, we integrate a Temporal Memory Fusion (TMF) module to retain critical information from the previous frames, which providing long-term contextual priors for the tracked objects.

We conduct experiments on the MOT17~\cite{dendorfer2021motchallenge} and DanceTrack~\cite{sun2022dancetrack} benchmarks to evaluate our proposed AR-MOT framework. Although our method does not surpass state-of-the-art approaches on several metrics, it represents the first attempt to address the MOT task within the LLM framework, offering greater potential for extension to more complex tasks. This work sets an important precedent for bridging MOT with general-purpose language models, and we believe it provides a solid foundation for future research.

Our contributions are summarized as follows:

\begin{itemize}
    \item We are the first to formulate the MOT task as an autoregressive sequence generation problem, enabling the model to support flexible formats and exhibit strong scalability.

    \item We propose object tokens, which further reformulate the MOT task as a multi-round dialogue process while enhancing the model’s fine-grained perceptual capability.

    \item We propose two novel modules, the Region-Aware Alignment (RAA) module and the Temporal Memory Fusion (TMF) module, which together enhance the model’s ability to maintain long-term target associations and achieve accurate feature alignment.
\end{itemize}

\section{Related Works}
\subsection{Tracking by Detection}
Tracking by Detection is a two-stage tracking paradigm that first detects objects in each frame and then associates them~\cite{bewley2016simple,wojke2017simple,du2023strongsort,cao2023observation,wang2020towards,zhang2021fairmot,zhang2022bytetrack,kong2022motfr,nodehi2021multi,you2021multi,deng2023jointing,jin2023multi,hu2023stdformer,jin2024ahor,shim2024enhancing,lin2024lttrack,liu2025sparsetrack,wang2022split,feng2023similarity,li2024single,zhang2023stat,zheng2024nettrack,jiang2025sam2mot,ravi2024sam,lv2024diffmot}. Recent studies focuses on associate through motion estimation and appearance features. SORT~\cite{bewley2016simple} is a classic model of TBD paradigm that using Kalman filtering for motion estimation and Hungarian Algorithm for object matching. DeepSORT~\cite{wojke2017simple} extends SORT by incorporating an extra appearance feature to improve inter-frame association. StrongSORT~\cite{du2023strongsort} further enhances DeepSORT by employing more powerful feature and a more comprehensive cost matrix to improve tracking performance. OC-SORT~\cite{cao2023observation} relaxes the linear motion assumption by estimating motion direction from bounding boxes to refine Kalman filter updates, thereby mitigating error accumulation and improving robustness under long-term occlusion or non-linear motion. To reduce dependence on standalone ReID networks, methods such as JDE~\cite{wang2020towards} and FairMOT~\cite{zhang2021fairmot} jointly optimize detection and identity embedding in a unified single-stage network. Unlike approaches that associate only high-confidence detections, ByteTrack~\cite{zhang2022bytetrack} first matches high-confidence detections to tracks, then associates low-confidence detections with unmatched tracks to retain potentially valid objects and suppress background noise, effectively handling occlusions. Recent works have integrated additional techniques into the tracking-by-detection framework. Nettrack~\cite{zheng2024nettrack} introduces point tracking to split the association process into box-level IoU matching and keypoint-to-box matching, enhancing robustness against deformation and rapid motion. SAM2MOT~\cite{jiang2025sam2mot} leverages the powerful segmentation model SAM2~\cite{ravi2024sam} to extend from single-object to multi-object tracking by directly generating tracking boxes from segmentation masks, reducing reliance on detection accuracy and enabling zero-shot generalization and strong association capabilities. DiffMOT~\cite{lv2024diffmot} adopts the core idea of diffusion models by formulating motion prediction as a denoising process conditioned on historical motion, achieving accurate tracking in non-linear motion scenarios via a decoupled, single-step sampling process.
\subsection{Tracking by Query}
In recent years, Tracking by Query has emerged as a novel paradigm in multi-object tracking. These methods utilize the query mechanism of Transformers to jointly perform object detection and cross-frame association in an end-to-end framework~\cite{sun2020transtrack,meinhardt2022trackformer,zeng2022motr,yan2023bridging,zhang2023motrv2,cai2022memot,gao2023memotr,segu2024samba}. A key component is the track query, which encodes spatial location and identity embedding of a tracked object and is iteratively updated over time to maintain temporal consistency. TransTrack~\cite{sun2020transtrack} is the first model to introduce Transformers into MOT. It employs learnable object queries to detect new objects and reuses features from the previous frame as tracking queries to achieve cross-frame association via query decoding. However, it still requires explicit matching between decoded detection boxes and predicted boxes.To eliminate explicit association, TrackFormer~\cite{meinhardt2022trackformer} and MOTR~\cite{zeng2022motr} adopt the DETR architecture. They update tracking queries frame by frame and detect new objects using object queries, achieving implicit association through the dynamic update of tracking queries. Although the MOTR framework is concise and effective, it presents challenges. CO-MOT~\cite{yan2023bridging} identifies potential conflicts between detection and association tasks and proposes assigning detection queries to match both new and tracked objects within intermediate decoder layers, thereby improving performance. MOTRv2~\cite{zhang2023motrv2} enhances detection by replacing initial detection queries with proposals generated from a strong pre-trained detector such as YOLOX~\cite{ge2021yolox}, providing detection priors for better accuracy. MeMOT~\cite{cai2022memot}, MeMOTR~\cite{gao2023memotr}, and SambaMOTR~\cite{segu2024samba} address the lack of long-term memory in this framework. By incorporating memory modules that fuse multi-frame historical features, they enable condition-based updates of track queries, significantly improving robustness under long-term occlusion.

\begin{figure*}[t]
\centering
\includegraphics[width=\linewidth]{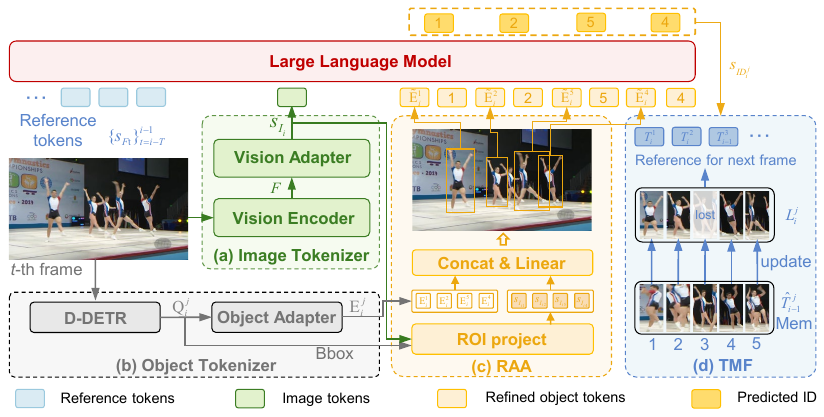}
\caption{\textbf{Framework of AR-MOT.} The proposed model consists of a Large Language Model along with two main modules (a)(b), and two additional modules (c)(d). (a) Image Tokenizer. Image Tokenizer is responsible for capturing global visual semantics from low-resolution images. It uses a SigLIP-ViT backbone to extract feature sequences, which are then projected into the LLM space through a vision adapter to produce image tokens. (b) Object Tokenizer. Object Tokenizer focuses on region-level perception by applying a DETR-based detector to high-resolution images. It generates object-specific features and projects them into the LLM space using an object adapter to obtain object tokens. (c) Region-Aware Alignment Module. RAA performs semantic alignment by projecting the bounding boxes from the Object Tokenizer onto the image tokenizer space to retrieve the corresponding image tokens, which are fused with object tokens to produce aligned object representations. (d) Temporal Memory Fusion Module. TMF maintains a memory of informative object tokens from earlier frames and dynamically integrates them into the current tokens. This design extends the temporal context while maintaining computational efficiency.}
\label{fig_2}
\end{figure*}

\subsection{Autoregressive Model}
With the advancement of large language models and sequence modeling techniques, researchers have begun exploring the feasibility of formulating vision tasks as sequence generation problems, leading to the emergence of autoregressive approaches in visual modeling. Pix2Seq~\cite{chen2021pix2seq} reformulates object detection as a sequence modeling task, enabling the model to autoregressively generate bounding boxes and class labels from an image. This design allows seamless integration with language models, offering a compact and elegant modeling paradigm. Pix2Seq-v2~\cite{chen2022unified} extends the framework to instance segmentation and multi-task settings, improving stability and accuracy through techniques such as data normalization and target encoding. Following the success of autoregressive modeling in traditional vision tasks, increasing attention has been directed toward applying such models to multimodal tasks. Gato~\cite{reed2022generalist}, OFA~\cite{wang2022ofa}, Florence-2~\cite{xiao2024florence}, and Unified-IO~\cite{lu2022unified} unify inputs and outputs of various vision-language tasks into token sequences and adopt autoregressive Transformer architectures to perform image captioning, object detection, segmentation, and visual question answering within a single framework. To further enhance language capabilities, recent multimodal large models incorporate pre-trained large language models to facilitate vision-language interaction. LLaVA~\cite{liu2023visual}, MiniGPT-4~\cite{zhu2023minigpt}, LLaVA-1.5~\cite{liu2024improved}, Otter~\cite{li2025otter}, and mPLUG-Owl~\cite{ye2023mplug} project visual features into the embedding space of language models and utilize staged pretraining and instruction tuning, demonstrating strong performance in vision-language dialogue and image understanding.

\section{Method}
In this paper, we propose AR-MOT, a novel framework that formulates the MOT task as an autoregressive sequence generation problem. In this section, we describe our autoregressive formulation for MOT and present the overall architecture of AR-MOT, along with details of its key components.

\subsection{Preliminary}
Autoregressive (AR) modeling has become a foundational paradigm for a wide range of sequence generation tasks. The core idea of an autoregressive model is to predict the next element in a sequence based on all previously generated elements, enabling the model to learn rich dependencies and structure over ordered data. Formally, given an input sequence $X = \{x_1, x_2, \cdots, x_n\}$, an autoregressive model aims to model the joint probability distribution as a product of conditional probabilities:
\begin{gather}
     P(x_1, x_2, \cdots, x_n) = \prod_{t = 1}^{n} P(x_t|x_1, x_2, \cdots, x_{t - 1})
\end{gather}
In practice, this formulation is commonly implemented using transformer-based decoder-only architectures, where the model predicts each token sequentially via next-token prediction. At each time step $t$, the model receives all previously generated tokens $\{x_1, x_2, \cdots, x_{t - 1}\}$ and outputs the probability distribution over the next token $x_t$.

Autoregressive models operate in a sequential manner, which supports customizable prompt design based on task requirements, offering flexible input formats and obviating the dedicated output heads, making autoregressive models inherently more extensible and easier to adapt across tasks. Besides, their effective sequence modeling capability ensures that outputs follow a well-defined structure, allowing the system to parse results directly without complex post-processing. Moreover, modeling MOT in an AR framework facilitates seamless integration with MLLMs, laying the groundwork for unified vision-language systems. These advantages motivate our formulation of the MOT task as a sequence generation problem under the autoregressive framework.

\subsection{Overview of AR-MOT}
Our proposed AR-MOT consists of three main components: a multimodal large language model (MLLM) of hidden dimension $d_{LM}$ equipped with a vision encoder of hidden dimension $d_S$, an object tokenizer equipped with an object detector of hidden dimension $d_D$, and a token alignment module. The overall architecture is illustrated in Fig. \ref{fig_2}.

Given a video sequence $S$ of length $M$, we formalize each frame as a set consisting of the image and its corresponding objects of interest. Each object of interest is further represented as a pair of an embedding and its corresponding ID. Formally, we formulate the video sequence with each frame as:
\begin{gather}
     S = \{F_i\}_{i = 1}^M \\
     F_i = \{I_i, \{\langle E^{j}_{i}, ID^{j}_{i} \rangle\}_{j=1}^{N_i}\}
\end{gather}
where $F_i$ represents the $i$-th frame, $I_i$ represents the image of the frame, $N_i$ represents the total number of objects in the frame, and $\langle E^{j}_{i}, ID^{j}_{i} \rangle$ represents the $j$-th object of the frame in the form of a pair of object embedding $E^{j}_{i}$ and corresponding ID $ID^{j}_{i}$.

Given image of $i$-th frame $I_i \in \mathbb{R}^{H \times W \times C}$, where $H$, $W$, and $C$ denote its height, width, and number of channels, respectively. To track objects in the $i$-th video frame, we first use a vision encoder to split the image into patches of size $P \times P$, extract visual features, and project them into the language model’s embedding space using a multilayer perceptron (MLP), resulting in an image token sequence $s_{I_i} \in \mathbb{R}^{(HW/P^2) \times d_{LM}}$. Next, the image $I_i$ is passed through an object tokenizer composed of an object detector D-DETR and an object adapter to get the embeddings with bounding boxes of all objects of interest in the frame. Each updated query $Q_{i}^{j} \in \mathbb{R}^{1 \times d_D}$ from D-DETR is then projected into the language model's embedding space, yielding an object token $E_{i}^{j} \in \mathbb{R}^{1 \times d_{LM}}$ that represents the $j$-th object.

On this basis, we model the MOT task as a sequence generation problem, in which the next token, representing the next object ID, is autoregressively predicted conditioned on the sequence of previously generated predictions. Therefore, given the past $T$ frames for reference to process $i$-th frame, the training objective of our model is defined as:
\begin{gather}
    \mathrm{maximize} \sum_{j=1}^{N_{i}} \mathrm{\log} P(s_{ID_i^j}|s_{past_i^j}) \\
    s_{past_i^j} = \mathrm{Concat}(\{s_{F_t}\}_{t=i-T}^{i-1},s_{I_i},\{s_{E_i^n}, s_{ID_i^n}\}_{n=0}^{j-1},s_{E_i^j})
\label{Eq. 5}
\end{gather}

where $P(\cdot)$ denotes the confidence score of predicting the next ID token, $s_{F_t}$ represents the historical information aggregated from all contents of the $t$-th frame. Under this paradigm, each embedding-ID pair can be viewed as a question-answer pair. The multi-object tracking task is carried out during the questioning process between the object detector and the LLM.

Multi-object tracking requires the ability to identify newly appeared objects. To address the emergence of new objects, we add a special token \texttt{<new>} into the vocabulary of the text tokenizer to indicate new objects. When predicting the first frame, or when the trajectory of a given ID has expired, the LLM predicts the object's ID as \texttt{<new>}. After completing the prediction for the current frame, an available ID is manually assigned to each new object.

The resulting sequence comprising image tokens, object tokens, and ID tokens is constructed as illustrated in \ref{fig_2} and then fed into an MLLM for autoregressively prediction.

\subsection{MLLM baseline}
We adopt the LLaVA-OneVision~\cite{li2024llava-onevision} model of the LLaVA-NeXT~\cite{li2024llava} framework as the backbone of the MLLM to build our AR-MOT. For the vision encoder, we employ a SigLIP~\cite{zhai2023sigmoid} image encoder with an input resolution of 384×384 to extract global visual features. These features are then projected into the language model’s space via an adapter layer composed of an MLP block. For the language model, we use the most lightweight model Qwen2~\cite{team2024qwen2} 0.5B to reduce the overall computational cost. 

\subsection{Object Tokenizer}
We introduce an object tokenizer that leverages object detectors to extract region-level features and encode them into LLM-compatible token representations to compensate for the lack of fine-grained perception in SigLIP-ViT. Given that detection transformer models inherently represent objects using a set of learnable queries and naturally aligns with our token-based formulation, we adopt Deformable DETR (D-DETR)~\cite{zhu2020deformable} as the object detector. We built AR-MOT based on D-DETR with ResNet-50~\cite{he2016deep} backbone to balance efficiency and convergence speed. The improved version of DETR offers strong performance and efficient multi-scale feature aggregation, which contribute to enhanced overall model capability. D-DETR detects a set of objects and outputs a corresponding set of object queries, encoding appearance and positional features relevant to specific regions. These queries are then transformed into LLM-compatible object tokens through an object adapter, which consists of a series of multi-layer perceptrons (MLPs).

\begin{figure}[!tbp]
\centering
\includegraphics[width=\linewidth]{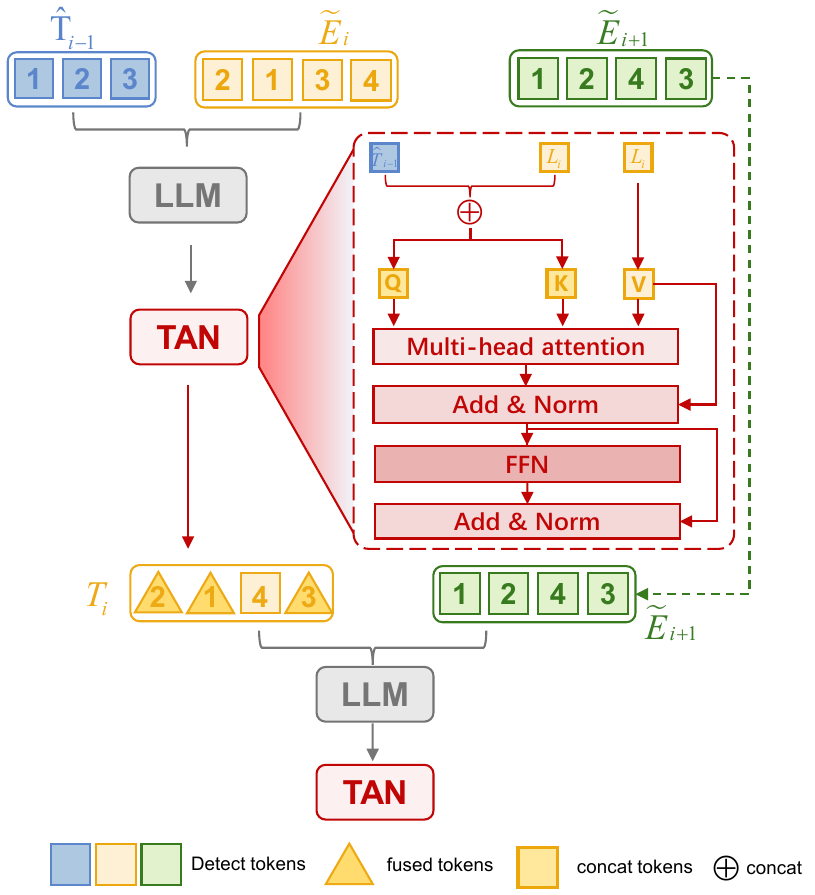}
\caption{\textbf{Temporal Memory Fusion module.} Upon completing tracking for the current frame, the corresponding object tokens are retained as reference for the next frame. The TMF module incorporates historical object tokens into the current ones, providing long-term temporal context for subsequent tracking.}
\label{fig_3}
\end{figure}

\subsection{Region-Aware Alignment}
In our model, image tokens and object tokens are designed to capture different aspects of visual information: the former encodes high-level semantic features, while the latter focuses on fine-grained local object details. These two types of tokens are extracted from separate visual encoders. However, due to this separation, directly feeding object tokens into the LLM can lead to semantic misalignment with the image tokens, resulting in suboptimal learning and convergence.

We introduce the Region-Aware Alignment (RAA) module to mitigate this issue, which explicitly aligns object tokens with their corresponding image-level semantics. Concretely, for detected objects, we first retrieve their bounding boxes from the detector. These bounding boxes are then projected into the space of the input image, allowing us to identify the sets of image patches they cover. We then extract the corresponding image tokens associated with bounding boxes and compute their element-wise average along the channel dimension to derive region-level semantic representations. Finally, we fuse these semantic tokens with the corresponding original object tokens using a linear layer and get aligned object tokens. Each aligned object token $\tilde{E}_{i}^{j}$ incorporates both local and global visual cues, facilitating better semantic consistency and improved convergence efficiency.

\subsection{Temporal Memory Fusion}
The historical information $s_{F_t}$ in Eq. \ref{Eq. 5} can not only be constructed by directly using a sequence of multiple historical frames, but also by compressing multi-frame information to reduce computational cost. Therefore, we design a Temporal Memory Fusion (TMF) module, which provides the aforementioned MOT system with an information compression option, thereby improving tracking efficiency. Inspired by the Temporal Aggregation Network (TAN) architecture empolyed in query-based frameworks such as MOTR~\cite{zeng2022motr}, we introduce the multi-head attention (MHA) layer to achieve temporal information aggregation and dynamic token updating. As illustrated in the Fig. \ref{fig_3}, upon completing the tracking process for the frame $I_i$, two primary inputs are fed into the TMF module: 1) the current frame hidden states $L_{i}^{j}$ generated by the MLLM, which contains the feature information of all objects in the current frame; 2) the historical frame embedding $\hat{T}^j_{i-1}$, which stores relevant features of the object in previous frames. Within the TMF module, these inputs are first fused to generate an enhanced feature representation. This representation is then used by the multi-head attention (MHA) mechanism as both queries (Q) and keys (K), while the embeddings of the current frame serve as values (V). This design enables dynamic aggregation of temporal information through attention-weighted contextualization. During the tracking of the frame $I_{i+1}$, the updated embedding $T_{i}^{j}$ id jointly  processed with the new frame embedding $\tilde{E}_{i+1}^{j}$ within the MLLM, forming continuously updated object representations that ensure temporal coherence and enhance tracking accuracy throughout the sequence.

\subsection{Training and Inference}
\subsubsection{Training}
During training, we simulate the tracking process by sampling video subsequences and training the model within these subsequences. Specifically, given a video sequence, we randomly sample a subsequence with temporal gap uniformly selected from the range $[0,L]$. Within this subsequence, we employ a sliding window approach to select a clip of $T + 1$ consecutive frames (where $2 \leq T + 1 \leq L$) as input for each training instance.

For each selected frame, all objects are first detected using the D-DETR model. The detection module is supervised by a combination of standard detection losses, including a classification loss $\mathcal{L}_{cls}$, an L1 regression loss $\mathcal{L}_{l1}$ and a generalized IoU loss $\mathcal{L}_{giou}$. Once detection is complete, the sequence is constructed as described in Section 3.2 and fed into the LLM. The predicted token sequence is then compared against the ground-truth token sequence using a cross-entropy loss $\mathcal{L}_{ce}$. The total training loss is thus formulated as:
$$\mathcal{L} = \lambda_{cls}\mathcal{L}_{cls} + \lambda_{l1}\mathcal{L}_{l1} + \lambda_{giou}\mathcal{L}_{giou} + \lambda_{ce}\mathcal{L}_{ce}$$
where $\lambda_{cls}$, $\lambda_{l1}$, $\lambda_{giou}$ and $\lambda_{ce}$ denote the weights for each corresponding loss.

\subsubsection{Inference Without TMF}
A sliding window is used to construct the reference history, consisting of the past $T$ frames. At each frame, bounding boxes are obtained by applying a confidence threshold $\tau_{det}$ to the detection model, and the corresponding updated queries are then projected and fed into the LLM to generate object IDs. 

To enhance the model's ability to recover lost objects, we maintain a Temporal Context Manager (TCM) that stores the latest object token for each object. This allows information from previously disappeared objects to be retained and incorporated into the reference sequence, even if those objects are currently absent. Such a design enables the model to recall occluded or temporarily exited objects and assign consistent IDs when they reappear. However, maintaining excessive trajectory history may introduce noise and ambiguity for LLM. Therefore, we implement a memory pruning strategy by tracking the number of frames $n_{lost}$ each object has been missing. A trajectory is considered expired and removed from the reference set when $n_{loss} > \tau_{loss}$, where $\tau_{loss}$ is a predefined threshold.

\subsubsection{Inference With TMF}
When TMF module is enabled, the sliding window is no longer used. Temporal information from previous frames is incrementally fused into a compact historical token sequence $T_{i-1}$. During inference, only $T_{i-1}$ and the current frame features are provided as reference to the LLM for ID generation.

\section{Experiments}
\subsection{Datasets and Metrics}
\subsubsection{Datasets}We selects the MOT17~\cite{dendorfer2021motchallenge} and DanceTrack~\cite{sun2022dancetrack} datasets to validate the effectiveness of the AR-MOT model. The MOT17 dataset contains 14 video sequences, with 7 sequences for training and 7 sequences for testing. The category of object in all sequences is pedestrians. The scenarios encompass both indoor and outdoor environments, captured from various camera viewpoints, under varying illumination conditions including daylight and nighttime. These characteristics make MOT17 a robust benchmark for evaluating detection and tracking performance in densely crowded scenes. In comparison, the DanceTrack dataset consists of 40 training, 25 validation, and 35 test video sequences. It focuses on human motion in fixed scenarios under complex lighting conditions, where objects exhibit more complex nonlinear motion patterns such as frequent occlusions, crossings, deformations, and also share highly similar visual appearances due to uniform costumes. This dataset provides a rigorous evaluation of a multi-object tracking model’s capability in motion modeling and a critical assessment of the model's performance on objects with extremely limited appearance discriminability.

\subsubsection{Metrics}We employ Clear MOT metrics~\cite{stiefelhagen2006clear} (e.g., MOTA), identification metrics~\cite{ristani2016performance} (e.g., IDF1), and higher-order metrics~\cite{luiten2021hota} (e.g., HOTA, DetA, AssA) to comprehensively evaluate the tracking performance of our model. MOTA provides an overall evaluation by jointly considering detection and association performance. However, it tends to place greater weight on detection errors while being less sensitive to association quality. IDF1 provides a comprehensive evaluation of a tracker's identity recognition and long-term identity preservation capabilities by jointly considering Identification Precision (IDP) and Identification Recall (IDR). This makes IDF1 a reliable measure of a tracker’s ability to maintain consistent identity in long-term tracking. HOTA further advances tracking evaluation by simultaneously capturing detection and association performance. It is defined as the geometric mean of DetA and AssA, ensuring a more balanced assessment and better reflects the model’s holistic tracking capability.

\subsection{Implementation Details}The model is initialized with official Deformable-DETR weights pretrained on COCO dataset~\cite{lin2014microsoft}. For each dataset, we perform a separate round of domain-specific pretraining on the corresponding training set (DanceTrack or MOT17) to ensure better adaptation to dataset-specific characteristics. For input processing, the shorter side of the input image is resized to 749, and the maximum size is restricted to 1333. All the experiments are conducted on PyTorch with a single NVIDIA RTX 4090 GPU with bfloat16 precision.

In our experiments, we set the random sampling interval for DanceTrack to range from 1 to 10. Specifically, we initialize the clip length as 2 frames and incrementally extend it to 3, 4, and 5 frames. All clips undergo comprehensive data augmentation including random cropping, multi-scale resizing, and horizontal flipping to enhance sample diversity. The model is trained for 15 epochs with a cosine annealing learning rate schedule, starting from an initial rate of $6.0 \times 10^{-5}$ using AdamW optimizer. This scheduling strategy effectively coordinates the learning rate decay with the progressive extension of temporal receptive fields.

\subsection{Ablation Study}
\subsubsection{Each Components of AR-MOT}
To evaluate the contribution of each proposed component to the overall performance of the AR-MOT framework, we conduct a series of incremental ablation studies on the MOT17 and DanceTrack datasets. We begin with a baseline model that retains only the MLLM component, where image features are extracted via the visual tokenizer and the LLM directly autoregressively predicts bounding boxes. Based on this setting, we progressively introduce the Object tokenizer, RAA and TMF modules to assess their individual impacts.

As shown in Table. \ref{tb:w/ w/o detr}, the results demonstrate that our autoregressive MOT formulation under the MLLM framework is capable of handling multi-object tracking tasks. In addition, the inclusion of each component significantly improves both detection and association performance.

\begin{table}[!tbp]
\centering
\caption{Effectiveness of the Object Tokenizer and Comparison Across Different Object Detectors.}
\label{tb:w/ w/o detr}
\begin{tabular}{c|ccccc} 
\toprule
  Method  & HOTA & DetA  & MOTA  &  AssA  &  IDF1    \\
  \midrule
 vanilla & 19.32 & 29.89 & 11.82 & 12.77 & 22.36   \\
 W/ DETR & \textbf{31.88} & \textbf{69.06} & \textbf{75.41} & \textbf{14.94} & \textbf{26.56}  \\
\bottomrule
\end{tabular}
\end{table}

To establish a baseline, we begin with a minimal configuration by retaining only the visual tokenizer and the large language model, removing all other modules such as object tokenizer. In this simplified setup, for each incoming video frame, only the image is provided as input to the model. All subsequent bounding boxes and IDs are generated in an auto-regressive manner. To facilitate the generation process, we introduce a special token \texttt{<next\_frame>} into the text tokenizer's vocabulary. During inference, the model generates bounding boxes one by one until it outputs the \texttt{<next\_frame>} token, which signals the end of the current frame. In our vanilla MLLM setup, the model achieved 19.32 HOTA and 11.82 MOTA on the DanceTrack val set. These results demonstrate that while SigLIP is capable of providing general semantic understanding of the input image, its lack of explicit pretraining for spatial awareness and object localization severely limits its ability to accurately detect and associate objects.

To address this issue, we introduce the object tokenizer into the framework. With this addition, the model's DetA score improved by 131 \%, and AssA increased by 17 \%. This performance gain highlights the effectiveness of explicitly modeling object-level tokens, which provide more structured and precise representations of visual objects. These enriched representations enable the model to better localize and associate objects across frames.

We further integrated the Temporal Memory Fusion module and Region-Aware Alignment module, which led to an additional improvement in overall performance. By jointly leveraging historical memory and region-level visual alignment, the model benefits from both temporal consistency and local perception.

\subsubsection{Analysis of Object Token}

\begin{table}[!tbp]
\centering
\caption{Comparison of Semantic and Positional Information in D-DETR's Query-Based Object Tokens vs. BBox-Encoded Object Tokens on dancetrack val set.}
\label{tb:different positional infomation}
\begin{tabular}{c|ccccc} 
\toprule 
  Method  & HOTA & DetA  & MOTA  &  AssA  &  IDF1    \\
  \midrule 
 box &  \textbf{31.88}  & \textbf{69.06}  & \textbf{75.41}  &  14.94  &  \textbf{26.56}  \\
 query &  31.19 &  65.19  &  69.47  &  \textbf{15.17}  &  26.32 \\
\bottomrule 
\end{tabular}
\end{table}

\begin{table}[!tbp]
\centering
\caption{Impact of Bounding Box Discretization Level.}
\label{tb:different coordinate discretization level}
\begin{tabular}{c|ccccc} 
\toprule 
  $\alpha$  & HOTA & DetA  & MOTA  &  AssA  &  IDF1    \\
  \midrule 
 0.4 &  28.97  & 60.16  & 63.96  &  14.43  &  25.08  \\
 0.5 &  31.67 &  63.90  &  70.00  &  16.11  &  27.17 \\
 0.6 & 33.05 & 67.49 & 73.62 & 16.51 & 28.09\\
 0.7 & 35.52  & 69.21 & 74.97 & 18.56 & 30.08\\
 0.8 & 35.81 & 70.54 & 76.86 & 18.46 & 30.41 \\
 1.0 & \textbf{37.09} & \textbf{71.64} & \textbf{78.01} & \textbf{19.48} & \textbf{31.67} \\
\bottomrule 
\end{tabular}
\end{table}

To compensate for the limited local localization capabilities of MLLMs, we incorporate an object detection model and introduce object tokens to provide the LLM with precise spatial information about visual objects. To explore how object tokens contribute to model performance, we conduct two comparative experiments under the same configuration. Firstly, we project the updated queries produced by the D-DETR detection model into the LLM space through an object adapter. Secondly, we replace object tokens with the raw bounding box coordinates of detection results. Specifically, we follow the encoding strategy from Pix2Seq~\cite{chen2021pix2seq}, discretizing the image space along the $x$ and $y$ axes into a total of $n_b$ bins. These discrete bin indices are added to the text tokenizer’s vocabulary as special tokens. The model takes the sequence of 4 tokens corresponding to a bounding box $[x_1, y_1, x_2, y_2]$ as an implicit prompt of the object’s location for ID generation.

As shown in Table. \ref{tb:different positional infomation}, box-encoded object tokens outperform query-based tokens, yielding 8.55\% higher MOTA and 5.94\% higher DetA. This improvement is mainly attributed to the explicit spatial and scale information encoded in bounding boxes, which provides more stable positional cues and facilitates more accurate object localization and association across frames.

As shown in Table. \ref{tb:different coordinate discretization level}, we investigate the impact of visual prompt granularity on the LLM’s tracking performance by varying the discretization level of the input bounding boxes. As the scaling factor $\alpha$ increases from 0.4 to 1.0, all metrics consistently improve, with MOTA rising from 63.96 to 78.01 and IDF1 from 25.08 to 31.67. This indicates that more precise visual information significantly enhances both detection and association, suggesting that the LLM benefits from relatively fine-grained perceptual cues provided by the visual backbone. These results highlight the importance of maintaining adequate visual resolution in prompts and motivate future work to explore strategies that balance token granularity with computational efficiency.

\subsubsection{Analysis of Region-Aware Alignment}

\begin{table}[!tbp]
    \centering
    \caption{Comparison of Different Model Components on dancetrack val set.}
    \label{tb:components eval}
    \begin{tabular}{ccc|ccccc}
    \toprule 
        MLLM & RAA & TMF & HOTA & DetA  & MOTA  &  AssA  &  IDF1 \\
         \midrule
        \checkmark  &  &  & 33.53 & 65.90 & 70.84 & 17.36 & 28.03 \\
        \checkmark & \checkmark &  & 32.59 & 71.57 & 77.84 & 15.08 & 26.02 \\
        \checkmark &  &  \checkmark & 33.74 & 71.05 & 76.83 & 16.22 & 29.17 \\
        \checkmark & \checkmark & \checkmark & \textbf{37.09} & \textbf{71.64} & \textbf{78.01} & \textbf{19.48} & \textbf{31.67} \\
    \bottomrule
    \end{tabular}
\end{table}

As shown in Table. ~\ref{tb:components eval}, both the RAA and TMF modules individually contribute to improvements in detection performance. Interestingly, when both modules are applied together, the model not only achieves the highest detection scores but also exhibits a substantial enhancement in association metrics. This suggests that while RAA and TMF each support detection, their combination enables more effective temporal and region-aware reasoning, which in turn significantly strengthens the model’s identity association capabilities in multi-object tracking.


\subsubsection{Analysis of Temporal Memory Fusion}
Our Temporal Memory Fusion (TMF) module propagates historical object information into the current reference sequence. To evaluate its effectiveness, we conduct experiments with and without the TMF module under identical settings. As shown in Table. \ref{tb:components eval}, the model achieves improvements of 5.99 \% in MOTA and 5.16 \% in DetA when using the TMF module. Under identical detection settings, the absence of the TMF module leads to ID mismatch for correctly detected objects consequently impacting detection-related metrics. The proposed TMF module enriches reference object tokens with long-term historical information, effectively reducing identity ambiguity. This enhancement improves ID prediction accuracy and increases the number of correctly identified object instances, contributing to better overall tracking performance.


\subsubsection{Analysis of Temporal Context Manager}

\begin{table}[!tbp]
\centering
\caption{Comparison of different $\tau_{loss}$ on dancetrack val set.}
\label{tb:different threshold's impact}
\begin{tabular}{c|ccccc} 
\toprule 
  $\tau_{loss}$  & HOTA & DetA  & MOTA  &  AssA  &  IDF1    \\
  \midrule 
 1 &  33.53  & 65.90  & 70.84  &  17.36  &  28.03  \\
 2 &  36.88 &  \textbf{70.75}  &  77.08  &  19.53  &  30.85 \\
 3 & 38.89 & 70.63 & 77.33 & 22.87 & 34.23 \\
 4 & 39.46 & 70.67 & 77.47 & 22.36 & 34.57 \\
 5 & 40.80 & 70.70 & 77.56 & 23.90 & 36.02 \\
 10 & 42.18 & 70.61 & \textbf{77.62} & 25.51 & 37.62 \\
 15 & \textbf{42.39} & 70.39 & 77.57 & \textbf{25.88} & \textbf{38.48} \\
\bottomrule 
\end{tabular}
\end{table}

\begin{figure}[!t]
\centering
\includegraphics[width=\linewidth]{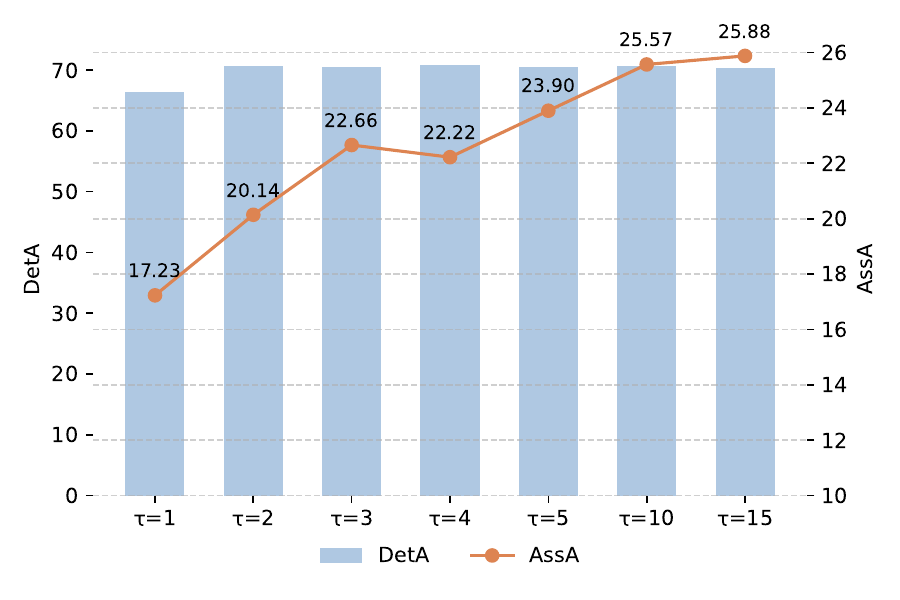}
\caption{\textbf{Impact of the $\tau_{loss}$ Parameter on DetA and AssA Metrics.} Preserving more historical frames with larger $\tau_{loss}$ has limited impact on DetA, while AssA changes more significantly, highlighting the importance of temporal history for association modeling.}
\label{fig_4}
\end{figure}

\begin{figure}[!t]
\centering
\includegraphics[width=\linewidth]{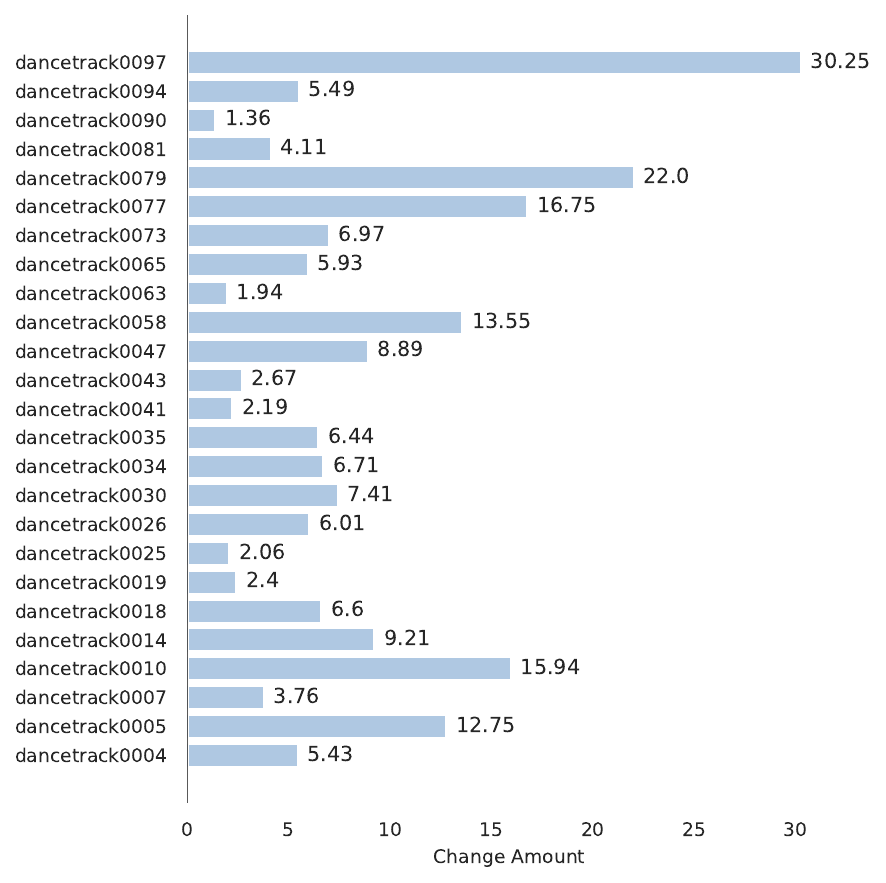}
\caption{\textbf{Sequence-wise Visualization Results at $\tau_{loss}=10$.} TCM boosts performance across all sequences, with notable enhancements on specific ones(e.g., dancetrack0097, dancetrack0079)}
\label{fig_5}
\end{figure}

We incorporate historical information stored in the Temporal Context Manager into the reference sequence during inference, allowing subsequent frames to retain access to these tokens. If a target remains undetected for more than $\tau_{loss}$ frames, the corresponding trajectory is discarded. As shown in Table. \ref{tb:different threshold's impact} and Fig. \ref{fig_4}, we evaluate the impact of different $\tau_{loss}$ thresholds on model performance. Increasing $\tau_{loss}$ allows the model to exploit longer temporal cues. While DetA shows only minor variations, AssA improves or fluctuates more significantly, suggesting that association accuracy relies heavily on historical information. When $\tau_{loss} = 1$, the model retains no historical information, leading to lower AssA scores.


To further analyze the effect of temporal memory retention under diverse conditions, we evaluate all 25 validation sequences in the DanceTrack dataset. As shown in Fig. \ref{fig_5}, temporal memory retention yields uneven performance gains across sequences. In particular, sequences involving long-term occlusions, irregular motion, or extended target absences benefit substantially from a larger temporal retention threshold ($\tau_{loss}$). In these cases, a higher threshold prevents premature identity removal, thereby reducing identity fragmentation and improving re-association after reappearance.


\subsubsection{Impact Analysis of MLLM}

To analyze the necessity of introducing MLLMs in MOT, we conduct ablation experiments by replacing the LLM decoder with a 6-layer Transformer causal decoder. As shown in Table. \ref{tb:different decoder}, this replacement leads to a significant drop in performance, with the model achieving only 5.24 AssA and 13.08 MOTA. Although accurate positional cues are provided, the model fails to utilize them effectively due to the lack of scene-level understanding. In contrast, using a pretrained LLM enables the model to interpret positional prompts correctly, resulting in a MOTA of 75.41 (+62.33), indicating that the LLM is essential for enabling effective tracking. This performance gap can be attributed to the pretrained LLM's ability to incorporate high-level contextual and semantic knowledge acquired during large-scale pretraining. Unlike the shallow Transformer decoder, which relies solely on positional embeddings and local token interactions, the LLM is capable of reasoning over spatial relationships, object continuity, and scene semantics. These capabilities allow it to resolve identity association and leverage positional cues more effectively, especially in ambiguous or occluded scenarios.

\begin{table}[!tbp]
\centering
\caption{Effectiveness of the different backbone.}
\label{tb:different decoder}
\begin{tabular}{c|ccccc} 
\toprule
  Causal decoder  & HOTA & DetA  & MOTA  &  AssA  &  IDF1    \\
  \midrule
 Transformer Decoder & 15.17 & 44.28 & 13.08 & 5.24 & 12.36   \\
 Qwen2-0.5B & \textbf{31.88} & 69.06 & \textbf{75.41} & \textbf{14.94} & \textbf{26.56}  \\
 Qwen2-7B (lora) & 29.36 & \textbf{69.70} & 74.52 & 12.57 & 24.01  \\
\bottomrule
\end{tabular}
\end{table}

\subsection{State-of-the-art Comparison}

\subsubsection{MOT17}
Existing MOT methods are commonly evaluated on the MOT17 benchmark using CLEAR metrics and IDF1 scores. Under this standard protocol, as shown in Table. \ref{tb:overall on mot17}, AR-MOT achieves a MOTA of 66.1 and an IDF1 of 60.7, verifying the effectiveness of the proposed autoregressive formulation in multi-object tracking. The slightly lower performance compared with state-of-the-art methods is mainly due to the large number of extremely small objects in MOT17, which are challenging for the visual encoder of the MLLM.

\subsubsection{DanceTrack}
AR-MOT exhibits strong performance on the DanceTrack dataset, where evaluation primarily focuses on the HOTA metric. As shown in Table. \ref{tb:overall on dancetrack}, without any post-processing and by directly generating object identities, our method outperforms the TBD-based DeepSORT by 2.5 HOTA and 0.5 AssA, despite DeepSORT employing a complex cascaded matching pipeline. It also exceeds ByteTrack, which utilizes a two-stage matching strategy based on detection confidence, by 0.1 HOTA. Compared with the query-based method TransTrack, ARMOT achieves significant gains of 2.6 HOTA and 2.7 AssA. Given the prevalence of occlusion and high appearance similarity in DanceTrack, these results highlight the robustness and effectiveness of our method in handling complex and visually challenging tracking scenarios.

\begin{table}[!tbp]
\centering
\caption{Comparison of AR-MOT with other methods on the MOT17 test set.}
\label{tb:overall on mot17}
\begin{tabular}{c|ccc} 
\toprule 
    Method  & MOTA$\uparrow$ & IDF1$\uparrow$  & HOTA$\uparrow$\\
    \midrule 
        MOTDT~\cite{chen2018real} & 50.9 & - & - \\
        DEFT~\cite{chaabane2021deft} & 66.6 & 65.42 & - \\
        CenterTrack~\cite{zhou2020tracking} & 67.8 & 61.4 & - \\
        QDTrack~\cite{fischer2023qdtrack} & 68.7 & 66.3 & - \\
        TraDeS~\cite{wu2021track} & 69.1 & 63.9 & - \\
        MOTR~\cite{zeng2022motr} & 73.4 & 68.6 & 57.8 \\
        FairMOT~\cite{zhang2021fairmot} & 73.7 & 72.3 & - \\
        MOTIP~\cite{gao2025multiple} & 75.3 & 71.3  & 59.2 \\
        GTR~\cite{zhou2022global} & 75.3 & 71.5 & 59.1 \\
        AHOR~\cite{jin2024ahor} & 82.3 & 80.7 & 65.6 \\
        LTTrack~\cite{lin2024lttrack} & 78.0 & 79.1 & 63.8 \\
        SparseTrack~\cite{liu2025sparsetrack} & 81.0 & 80.1 & 65.1 \\
        \textbf{AR-MOT(ours)} & 66.1 & 60.7 & 50.1 \\
\bottomrule 
\end{tabular}
\end{table}

\begin{table}[!tbp]
\centering
\caption{Comparison of AR-MOT with other methods on the dancetrack test set.}
\label{tb:overall on dancetrack}
\begin{tabular}{c|ccccc} 
\toprule 
    Method  & HOTA$\uparrow$ & DetA$\uparrow$  & MOTA$\uparrow$  &  AssA$\uparrow$  &  IDF1$\uparrow$\\
    \midrule 
        FairMOT~\cite{zhang2021fairmot} & 39.7 & 66.7 & 82.2 & 23.8 & 40.8\\
        CenterTrack~\cite{zhou2020tracking} & 41.8 & 78.1 & 86.8 & 22.6 & 35.7\\
        TraDeS~\cite{wu2021track} & 43.3 & 74.5 & 86.2 & 25.4 & 41.2\\
        TransTrack~\cite{sun2020transtrack} & 45.5 & 75.9 & 88.4 & 27.5 & 45.2\\
        QDTrack~\cite{fischer2023qdtrack} & 45.7 & 72.1 & 83.0 & 29.2 & 44.8\\
        DeepSORT~\cite{wojke2017simple} & 45.6 & 71.0 & 87.8 & 29.7 & 47.9\\
        ByteTrack~\cite{zhang2022bytetrack} & 47.7 & 71.0 & 89.6 & 32.1 & 53.9\\
        GTR~\cite{zhou2022global} & 48.0 & 72.5 & 84.7 & 31.9 & 50.3\\
        MOTR~\cite{zeng2022motr} & 54.2 & 73.5 & 79.7 & 40.2 & 51.5 \\
        MOTIP~\cite{gao2025multiple} & 69.6 & 80.4 & 90.6 & 60.4 & 74.7 \\
        DiffusionTrack~\cite{luo2024diffusiontrack} & 52.4 & 82.2 & 89.3 & 33.5 & 47.5 \\
        AHOR~\cite{jin2024ahor} & 60.2 & 82.2 & 91.4 & 44.2 & 59.0 \\
        LTTrack~\cite{lin2024lttrack} & 58.8 & 81.1 & - & 43.0 & 60.5 \\
        SparseTrack~\cite{liu2025sparsetrack} & 55.5 & 78.9 & 91.3 & 39.1 & 58.3 \\
        \textbf{AR-MOT(ours)} & 48.1 & 77.2 & 86.3 & 30.2 & 43.9\\
\bottomrule 
\end{tabular}
\end{table}

\subsection{Discussion}
AR-MOT formulates the multi-object tracking (MOT) task as an autoregressive prediction problem. Experimental results demonstrate that it is capable of effectively performing MOT and achieving competitive performance. However, several challenges remain to be addressed. Since ARMOT is derived from the tracking-by-detection (TBD) paradigm, it requires the joint optimization of both the detection model and the LLM using different loss functions, which introduces additional complexity during training. In addition, the overall tracking performance of the model is highly dependent on the quality of the detection results.

\section{Conclusion}
To make the multi-object tracking (MOT) task compatible with large language model (LLM) frameworks and to expand its application scope, we propose AR-MOT, the first approach that formulates MOT as an autoregressive sequence generation problem.  AR-MOT represents all essential elements including image features, object queries, and ID embeddings, as a unified sequence of tokens, and performs tracking by generating these tokens in an autoregressive manner. To further enhance the representational capacity of object tokens, we incorporate rich appearance features through Region-Aware Alignment (RAA) module and Temporal Memory Fusion (TMF) module. With this design, existing LLMs can be adapted for multi-object tracking through just fine-tuning, without requiring task-specific architectural modifications. We hope AR-MOT will encourage further research toward more generalized and scalable MOT paradigms and facilitate the application of MOT in a wider range of scenarios.

\bibliographystyle{IEEEtran}
\bibliography{references}{}


 




\vfill

\end{document}